\documentclass{article}

\usepackage{arxiv}

\usepackage[utf8]{inputenc} 
\usepackage[T1]{fontenc}    
\usepackage{hyperref}       
\usepackage{url}            
\usepackage{booktabs}       
\usepackage{amsfonts}       
\usepackage{nicefrac}       
\usepackage{microtype}      
\usepackage{lipsum}
\usepackage{titlesec}
\usepackage{amsmath}
\usepackage{graphicx}
\usepackage{verbatim}
\usepackage[]{quoting}

\titleformat{\paragraph}
{\normalfont\normalsize\bfseries}{\theparagraph}{1em}{}
\titlespacing*{\paragraph}
{0pt}{3.25ex plus 1ex minus .2ex}{1.5ex plus .2ex}

\title{Features in Extractive Supervised Single-document Summarization: Case of Persian News}

\author{
  Hosein~Rezaei \\
  Department of Electrical and Computer Engineering\\
  Isfahan University of Technology\\
  \texttt{hosein.rezaei@alumni.iut.ac.ir} \\
   \And
 Seyed Amid MoeinZadeh \\
  Department of Electrical and Computer Engineering\\
  Isfahan University of Technology\\
  \texttt{Amid.moeinzadeh@gmail.com} \\
   \AND
  Azar Shahgholian \\
   Liverpool John Moores University \\
   Liverpool Business School \\
   \texttt{A.Shahgholian@ljmu.ac.uk} \\
   \And
   Mohamad Saraee \\
   School of Computing, Science \& Engineering  \\
   University of Salford \\
   \texttt{M.saraee@salford.ac.uk} 
}

\begin{document}
\maketitle

\begin{abstract}
Text summarization has been one of the most challenging areas of research in NLP. Much effort has been made to overcome this challenge by using either the abstractive or extractive methods. Extractive methods are more popular, due to their simplicity compared with the more elaborate abstractive methods. In extractive approaches, the system will not generate sentences. Instead, it learns how to score sentences within the text by using some textual features and subsequently selecting those with the highest-rank. Therefore, the core objective is ranking and it highly depends on the document. This dependency has been unnoticed by many state-of-the-art solutions. In this work, the features of the document are integrated into vectors of every sentence. In this way, the system becomes informed about the context, increases the precision of the learned model and consequently produces comprehensive and brief summaries.
\end{abstract}


\section{Introduction}
From the early days of artificial intelligence, automatically summarizing a text was an interesting task for many researchers. Followed by the advance of the World Wide Web and the advent of concepts such as Social networks, Big Data, and Cloud computing among others, text summarization became a crucial task in many applications \cite{Mishra_Text_2014}, \cite{Sakai_Generic_2001}, \cite{Buenaga_Mana_Lopez_2004}. For example, it is essential, in many search engines and text retrieval systems to display a portion of each result entry which is representative of the whole text \cite{Roussinov_Information_2001}, \cite{Turpin_Fast_2007}. It is also becoming essential for managers and the general public to gain the gist of news and articles immediately, in order to save time, while being inundated with information on social media \cite{McKeown2005}.

Researchers have approached this challenge from various perspectives and have obtained some promising results \cite{ferreira2014context}, \cite{Barrera2012}. However, this area continues to present more research challenges and has a long path to maturity. 

One method of investigating this challenge, is (supervised) extractive summarization. Extractive implementations use a ranking mechanism and select top-n-ranked sentences as the summary \cite{Gupta2010}. Sentences of a document are represented as vectors of features. Using summarization corpora, a rank will be assigned to each sentence, based on its presence in several human-written summaries (golden summaries). The system should then learn how to use those features to predict the rank of sentences in any given text. Various machine learning approaches such as regression and classification algorithms are used to perform the ranking task \cite{Wong2008}, \cite{Hirao_Extracting_2002}. 

As far as our knowledge goes, in all current implementations, sets of sentence vectors of every document are merged together to compose a larger set, which is then passed to the learning model as a matrix. In this approach, the locality of ranks is disregarded. In other words, the rank of sentences is highly relative to the context and document. A sentence might be ranked high in one document while being ranked lower in another. As a result, merging sentences of a whole dataset into a matrix removes document boundaries and a main source of information will be lost. 

We addressed this issue by taking certain features of documents into account, such as its length, topical category and so on in addition to some new sentence features that also reflect document properties. Thus, more information will be provided to the model, and ranking could be done with respect to local features of the document. Our experiments show that this rectification leads to improvement in both the performance of the learned model and the quality of produced summaries. 
We also represent a new baseline for the evaluation of extractive text summarizers which can be used to measure the performance of any summarizing method more accurately. 

The remainder of this paper is organized as follows. (Section~\ref{sec:relateds}) reviews related works. (Section~\ref{sec:methodology}) presents the proposed method and evaluation measures. (Section~\ref{sec:results}) discusses how the experiments are set up. The results are discussed in (Section~\ref{sec:results}), and finally (Section~\ref{sec:conclusion}) concludes the paper.

\section{Related works}
\label{sec:relateds}

Text summarization has been widely studied by both academic and enterprise disciplines. Text summarization methods may be classified into different types. Based on input type, there are single-document \cite{Torres-Moreno_M._2014}, \cite{Patil_Automatic_2015} vs multi-document summarization methods \cite{Christensen_Towards_2013}, \cite{Erkan2004}, \cite{Nenkova_A_2006}. Based on language, there are mono-lingual, bilingual and multi-lingual methods \cite{Gambhir_Recent_2017}. There are also “query focused” methods in which a summary relevant to a given query is produced \cite{Varadarajan2006}. From the perspective of procedure, however, there are two main approaches: abstractive vs extractive \cite{Hahn_The_2000}. 

Abstractive approaches try to generate a new short text based on the concepts understood from the original text \cite{Moratanch_A_2016}. This usually requires a full pass through NLP pipeline and is faced with many complexities and challenges \cite{Lloret2012}. The abstractive approach relies on linguistic methods to examine and interpret the text in order to find new concepts and expressions. The output is a new shorter text which consists of the most important information from the original text document \cite{Gupta2010}. 

Extractive approaches, on the other hand, select a few sentences from the document based on some measures in order to place them in a summary \cite{Gupta2010}.

A broad range of methods has been examined in this approach, including graph-based \cite{Gupta2010}, \cite{Mihalcea}, unsupervised \cite{Mihalcea}, \cite{Rautray_An_2017} and supervised (corpus-based) methods \cite{Wong2008}, \cite{Silva_Automatic_2015}, \cite{Shafiee_Similarity_2018}. In supervised methods, training data is generally needed to select important content from the documents. In these methods, usually, the problem is reduced to a classification or regression problem, and machine learning techniques applied to the dataset of documents and their gold summaries represented by some features. Support Vector Machines (SVM) \cite{Ouyang_Applying_2011} and neural networks \cite{Fattah_A_2014} are more popular sentence classification algorithms.

The key step in extractive summarization is to determine the importance of sentences in the document \cite{Fang_Word-sentence_2017}. Previous studies examine the ordinal position of sentences \cite{Edmundson_New_1969}, \cite{Fattah_Automatic_2008}, length of sentences \cite{Wong2008}, the ratio of nouns, the Ratio of Verbs, Ratio of Adjectives, Ratio of Adverbs \cite{Dlikman_Using_2016}, the Ratio of Numerical entities \cite{Ferreira_Assessing_2013}, \cite{Lin_Y._1999} and Cue Words \cite{Edmundson_New_1969}.

Gupta and Lehal in their survey of text summarization techniques list the following groups of features: content-based, title-based, location-based, length-based, proper noun and upper-case word-based, font-based, specific phrase-based, and features based on sentence similarity to other sentences in a text \cite{Gupta2010}. Previous studies use different sentence features such as terms from keywords/key phrases, terms from user queries, frequency of words, and position of words/sentences for text summarization \cite{Ozsoy_Text_2011}.

However, in most cases, selection and weighting of features are an important matter of debate. Some works have been carried out with respect to this \cite{Neto2002}, but none, to the best of our knowledge, has shown that target attribute is highly related to the scope of the document. It is occasionally mentioned but not included in practice. For instance, Ferreira et al studied various combinations of sentence scoring methods on three types of documents in \cite{ferreira2014context} and \cite{Ferreira_Assessing_2013} and concluded that the weight of features varies, dependent on the properties of context: \begin{quoting}
“the effectiveness of sentence scoring methods for automatic extractive text summarization algorithms depends on the kind of text one wants to summarize, the length of documents, the kind of language used, and their structure.”.\end{quoting}
 JY Yeh et al in \cite{Yeh_Y_2005} utilized a Genetic Algorithm (GA) to find the weight of features for calculating sentence scores. However, their following statement implies that performance of weights is generally dependent to genre, that could be seen as a feature of context: \begin{quoting}“It cannot be guaranteed that the score function whose feature weights are obtained by GA definitely performs well for the test corpus; nevertheless, if the genre of the test corpus is close to that of the training corpus, we can make a prediction that the score function will work well.”\end{quoting} \cite{Yeh_Y_2005}. Berenjkoub et al studied the effectiveness of various subsets of features in summarization of distinct sections of scientific papers \cite{Berenjkoub_Persian_2012}. They showed that some features work well only in some specific portion of text, for example, on the abstract section, while others perform better on the methodology section. This could be considered to be a consequence of differences in the structure and context of each section. 

All the above studies imply the significance of document context in ranking. Nevertheless, it has not been given enough attention in the NLP community, and even sometimes is neglected. For instance, authors in \cite{Dlikman_Using_2016} suggest the use of a wide range of various features. Among these, seventeen part-of-speech based sentences features have been introduced, all of which are sentence-normalized, but not document-normalized, i.e. they count the ratio of a syntactic unit e.g. verbs, divided by the number of words in a sentence. Such features do not consider the total number of those units, e.g. verbs, in the whole document. 

 Our work contributes to this line of research and includes document features in the learning and ranking processes.

\section{Incorporating Document Features}
\label{sec:methodology}
As a way to investigate the need for document features in sentence ranking (as explained in the introduction and related works), we introduced several document-level features and incorporated them in the summarization process. These features are listed under subsection (\ref{subsec:learning-FE}). Although stages of our method do not differ from general supervised extractive summarization, the whole process is explained in order to clarify the method of investigation.

Every supervised summarization has two phases. The first is the “Learning Phase”, a corpus of ideal summaries is used to train the system how to rank sentences. The second is the “Summarization Phase”, where the system applies its learning gained from the first phase, in order to rank the sentences of a new given text. A process of selection is then performed to form a summary. Each of these phases has several intricacies which are briefly described in the following sections.

\subsection{Learning Phase}
The input to this phase is a dataset of documents, each of which is associated with several human-written summaries. The output is a learned model with a good level of accuracy that is able to reliably predict the rank of sentences, in almost the same way that a human may rank them. To accomplish this, it is necessary to first perform normalization and transform various forms of phrases into their canonical form. Then, every text should be tokenized to sentences, and further tokenized to words. Another prerequisite is to remove stop words. The following subtasks should be carried out next.

\subsubsection{Feature Extraction}
\label{subsec:learning-FE}
Foremost, it is necessary to represent each sentence with those features that have the most distinguishing effect on the prediction of the rank. Many features have been examined in the literature. We entitle some as “document-aware” because they do implicitly represent some information about a document. However, other features have been used, that say nothing about the document in which they appeared. We call them “document-unaware”. In the previous sections, we argued that this lack of information might be misleading for the system, especially when we train it with sample sentences from different documents. Thus, we modified some document-unaware features and derived new features that cover document properties. We also examined the effect of incorporating explicit features of a document into vectors of its sentences. The following sub-sections describe the features mentioned above in more detail.

\paragraph{Document-unaware Features}
\label{subsec:unawares}
\textbf{Ordinal position:} It is shown that inclusion of sentence, in summary, is relevant to its position in the document or even in a paragraph. Intuitively, sentences at the beginning or the end of a text are more likely to be included in the summary. Depending on how it is defined, this feature might be document-unaware or not. For example, in \cite{Fattah_Automatic_2008} and \cite{Suanmali_Fuzzy_2009} it is defined as $\frac{5}{5}$ for the first sentence, $\frac{4}{5}$ for the second, and so on to $\frac{1}{5}$ for fifth and zero for remaining sentences. In another research conducted by Wong et al. \cite{Wong2008}, it is defined as $\frac{1}{sentence\ number}$. With such a definition, we may have several sentences, for example, with position=$\frac{1}{5}$ in the training set,  these may not have the same sense of position. While a sentence position=$\frac{1}{5}$ means “among the firsts” in a document with 40 sentences, it has a totally different meaning of “in the middle”, in another document containing 10 sentences. Thus, a useful feature formula should involve differences of documents which may change the meaning of information within it. In our experiments, we used the definition of \cite{Wong2008}. A document-aware version of position will be introduced in (\ref{parag:awares}).

\textbf{Length of sentence:} the intuition behind this feature is that sentences of too long or too short length are less likely to be included in the summary. Like sentence position, this feature is also subject to the wrong definition that makes it document-unaware. For example, in \cite{Wong2008} it is defined as a number of words in a sentence. Such a definition does not take into account that a sentence with, say 15 words may be considered long if all other sentences of document have fewer words. Another sentence with the same number of words may be regarded as short, because other sentences in that document have more than 15 words. This might occur due to different writing styles. However, we included this in our experiments to compare its effect with that of its document-aware counterpart, which will be listed in (\ref{parag:awares}).

\textbf{The Ratio of Nouns:} is defined in \cite{Dlikman_Using_2016} as the number of nouns divided by total number of words in the sentence, after stop-words are removed. Three other features, Ratio of Verbs, Ratio of Adjectives, and Ratio of Adverbs are defined in the same manner and proved to have a positive effect on ranking performance. From our perspective, however, a sentence with a ratio of nouns =0.5, for example, in a document containing many nouns, must be discriminated in the training set from another sentence with the same ratio of nouns, that appeared in another document having fewer nouns. This feature does not represent how many nouns are there in the document, which is important in sentence ranking. The same discussion goes on to justify the need to consider the number of verbs, adjectives, and adverbs in the document. The impact of these features is examined in our experiments and compared to that of their document-aware counterparts. 

\textbf{The Ratio of Numerical entities:} assuming that sentences containing more numerical data are probably giving us more information, this feature may help us in ranking \cite{Ferreira_Assessing_2013}, \cite{Lin_Y._1999}. For calculation, we count the occurrences of numbers and digits proportional to the length of sentence. This feature must be less weighted if almost all sentences of a document have numerical data. However, it does not count numbers and digits in other sentences of the document.

\textbf{Cue Words:} if a sentence contains special phrases such as “in conclusion”, “overall”, “to summarize”, “in a nutshell” and so forth, its selection as a part of the summary is more probable than others. The number of these phrases is counted for this feature.

\paragraph{Document-aware Features}
\label{parag:awares}
\textbf{Cosine position:} As mentioned in (\ref{subsec:unawares}) a good definition of position should take into account document length. A well-known formula used in the literature \cite{Verma_Document_2010}, \cite{Barrera2012} is 

\begin{equation}
\label{eq:pos}
    pos(index) = \frac{\cos(\frac{2\pi\times index}{T-1})+\alpha -1}{\alpha}
\end{equation}

in which index is an integer representing the order of sentences and T is the total number of sentences in document. This feature ranges from 0 to 1, the closer to the beginning or to the end, the higher value this feature will take. $\alpha$ is a tuning parameter. As it increases, the value of this feature will be distributed more equally over sentences. In this manner, equal values of this feature in the training set represent a uniform notion of position in a document, so it becomes document-aware.

\textbf{Relative Length:} the intuition behind this feature is explained in (\ref{subsec:unawares}). A discussion went there that a simple count of words does not take into account that a sentence with a certain number of words may be considered long or short, based on the other sentences appeared the document. Taking this into consideration, we divided the number of words in the sentence by the average length of sentences in the document. More formally, the formula is:

\begin{equation}
Relative\ Length(s)=  \frac{|s|}{\frac{\Sigma _1^n|s_i |}{n}} 
\end{equation}
  
in which n is number of sentences in the document and $s_i$ is the i’th sentence of it. Values greater than 1 could be interpreted as long and vice versa.

\textbf{TF-ISF:} this feature counts the frequency of terms in a document and assigns higher values to sentences having more frequent terms. It also discounts terms which appear in more sentences. Since it is well explained in the literature, we have not included details and formula which are in references \cite{Neto2002} and \cite{Neto2000}. Nonetheless, the aspect that matters in our discussion is that both frequency and inverse sentence frequency are terms which involve properties of context, and consequently are document-aware.

\textbf{POS features:} Here we introduce another way to include the ratio of part of speech (POS) units in features and keep them document-normalized. To do this, the number of occurrences of each POS unit should be divided by the number of them in the document, instead of that occurring in a sentence.
The formal definition of the new document-aware features are as follows:

\begin{equation}
Ratio\ of\ nouns\ in\ document(s) = \frac{number\ of\ nouns\ in\ s}{number\ of\ nouns\ in\ document}
\end{equation}
\begin{equation}
Ratio\ of\ verbs\ in\ document(s) = \frac{number\ of\ verbs\ in\ s}{number\ of\ verbs\ in\ document}
\end{equation}
\begin{equation}
Ratio\ of\ adjectives\ in\ document(s) = \frac{number\ of\ adjectives\ in\ s}{number\ of\ adjectives\ in\ document}
\end{equation}
\begin{equation}
Ratio\ of\ adverbs\ in\ document(s) = \frac{number\ of\ adverbs\ in\ s}{number\ of\ adverbs\ in\ document}
\end{equation}
\begin{equation}
Ratio\ of\ numbers\ in\ document(s) = \frac{number\ of\ numerical\ entities\ in\ s}{number\ of\ numerical\ entities\ in\ document}
\end{equation}

\paragraph{Explicit Document Features}
In order to further investigate how effective are document specific features in sentence ranking, we defined several features for documents. These features are then calculated for each document and repeated in the feature vector of every sentence of that document. Their formal definition is described below and their effect is examined in the result and discussion section (\ref{sec:results}):

\textbf{Document sentences:} An important property of a document that affects summarization is the total number of sentences participating in sentence ranking. As this number grows, a summarizer should be more selective and precise. Also, some sentence features such as cue words, maybe more weighted for longer documents. In addition, the main contextual information is probably more distributed over sentences. In such a case even lower values of other features should be considered important.

\textbf{Document words:} the number of words in the document is another notion of document length. Since the number of sentences alone is not enough to represent document length, this feature should also be considered.

\textbf{Topical category:} different topics such as political, economic, etc. have different writing styles and this might affect sentence ranking. For instance, numerical entities may appear more in economic or sport reports than in religious or social news. Therefore the weight of this attribute should be more or less, based on a document’s category. So it needs to be included.

\begin{figure}
  \centering
  \includegraphics[width=0.7\linewidth]{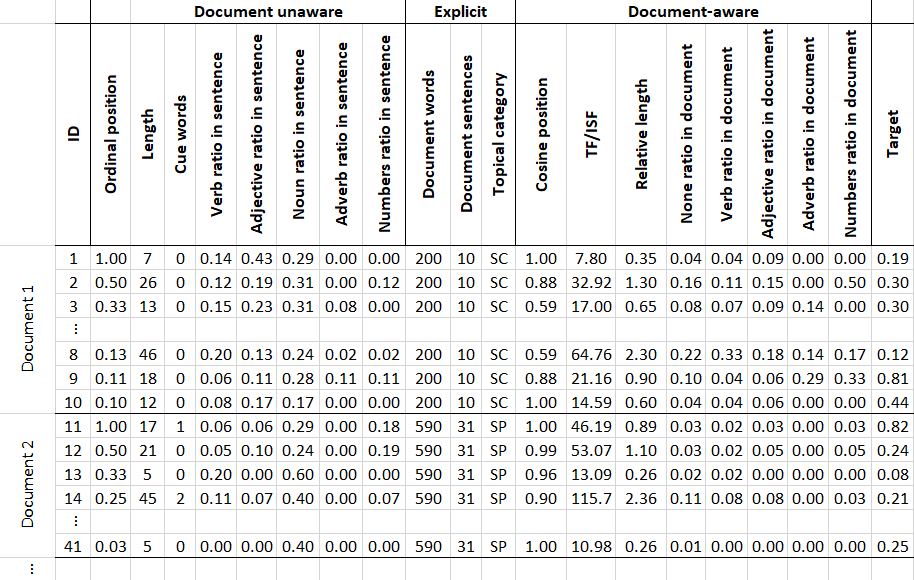}
  \caption{An excerpt of whole feature set. SC and SP under Topical category stand for Science and Sport, respectively.}
  \label{fig:fig1}
\end{figure}

An overview of our feature set is represented by example in figure \ref{fig:fig1}. Column ID is just for enumeration and column Target is explained in the next section.

\subsubsection{Target Assignment}
\label{sec:learning-TA}
Every feature vector needs a target value from which the system should learn how to rank sentences. The value of target is usually determined based on golden summaries. If a sentence is included in a majority of human-written extracts, its target is near to 1. In contrast, it would be closer to 0 if the sentence could not be found in any human-made summaries. In some datasets, like the one we used, golden summaries are not absolutely extractive, and they are not composed of exact copies of sentences in the original text. In such cases, a measure of similarity between the sentence whose target we are looking for, and each ideal summaries’ sentence will be calculated. This results in real values between 0 and 1 for this attribute. Section (\ref{sec:experiments}) includes more details about target assignment. 
\subsubsection{Training Model}
Since target attribute values vary between zero and one, we opted to use regression methods for the learning task. To build a training and a test set, a global matrix is composed in which every row corresponds to a sentence in the corpus and each column corresponds to a feature. The last column is for target attribute which will be omitted in the test set. It might be required to perform scaling on certain columns, depending on its corresponding feature and range of values.

In cases where the dataset is large, the total number of sentences which are not included in golden summaries, and consequently have lower targets, is many times larger than the number of included sentences. This might lead the regression bias toward lower target values. To avoid this, dataset balancing is needed. That is to leave aside a portion of not included sentences and not to feed them to learner model. 

Lastly, in this phase, the regression model should be fitted on training set and be evaluated on a test set as described in sections (\ref{sec:experiments}) and (\ref{sec:results}). 
\subsection{Summarization Phase}
Having acquired a model that can precisely rank sentences, we can apply it to any new given text and use ranked sentences in order to create a summary. This summarization process could also be executed on dataset texts, in order to evaluate how precisely our method resembles human-written summaries. In this section, we briefly describe the summarization process. The evaluation process is explained in section (\ref{sec:eval}).
\subsubsection{Feature Extraction}
Initially, sentence features need to be extracted. Again, normalization, sentence tokenization, word tokenization, and stop words removal are preliminary steps. The same features used in the learning phase should be calculated. 
\subsubsection{Sentence Ranking}
In comparison with learning phase, in which a global matrix was used, this time a local matrix is composed whose rows correspond with the sentences of the input text. If during learning, any scaling was performed on features, they should be carried out here in the same manner. The matrix is then fed to the regressor obtained in the previous phase, and a rank value between zero and one will be predicted for each sentence.
\subsubsection{Sentence Selection}
By sorting sentences based on their ranks, the most appropriate sentences for being included in summary will be determined. To preserve readability, however, it is important to place them in the summary in the same order they appeared in the input document. 

Another consideration is the cut-off length. How many of the top sentences should we select for summary? The answer should be as simple as a constant number, a percentage of total sentences, or it could be determined by more advanced heuristics. We allowed cut-off length to be an input parameter. This allows us, in the evaluation phase, to produce summaries of dataset documents in the same length as golden summaries. This makes the comparison more equitable. 
\subsection{Evaluation Measures}
\label{sec:eval}
In this section, some measures are described to evaluate the performance of both phases explained in the previous section: the learning phase and summarization phase. The former is evaluated using common regression metrics such as mean square error (MSE) and coefficient of determination (R2). The latter is carried out using ROUGE which is a well-known metric for evaluating summarization systems. 

Mean Square Error (MSE) is the average of squared errors in all estimated targets. An ideal regressor tends to make this measure as near as possible to zero. Though, an exact zero for MSE is not desirable, because it is suspected to be due to over fitting. 

The coefficient of determination is another metric for evaluating how well a regression model is fitted to data. It ranges from $-\infty$ to 1. As it approaches 1, “goodness-of-fit” is increased, while negative values show that the mean of data is a better estimator for target \cite{Nagelkerke_A_1991}.

ROUGE is proposed in \cite{Lin_Y._2004} as an evaluation metric for summaries. It matches n-grams in both system produced summaries and reference summaries and returns the percentage of matching in terms of precision, recall and f-measure. There is a variety of ROUGE family metrics, namely ROUGE-1, ROUGE-2, and ROUGE-L. In ROUGE-1 the overlap of 1-grams, each word, is calculated. In ROUGE-2 the bigrams are considered as units of comparison. The ROUGE-L uses the Longest Common Subsequence (LCS) to measure resemblance. Nevertheless, we found that ROUGE assessments are always relatively high, even for a summary that is produced perfunctorily. Hence, we also designed a random summarizer that selects random sentences for the summary, and evaluated it by ROUGE. This could be used as a baseline for comparison. 
\section{Experiments}
\label{sec:experiments}
Two experiments were set up to verify our hypothesis: “sentence ranking is highly dependent to document, and features must also represent context”. The first experiment involves document-unaware features (listed in section \ref{subsec:unawares}) alongside TF-ISF. In the second experiment, document-aware features were used instead of document-unaware ones. We also set up a random summarizer based on a random regressor that acts as a baseline for comparisons. More details are recorded in section (\ref{sec:expr-TA}).

A good experimental study should be as reproducible as possible. Here we explain the technical details that are more specific to our dataset, to allow the interested user to set up the same experiments for further research.
\subsection{Dataset}
We used the Pasokh dataset \cite{Moghaddas_Pasokh:_2013} that contains 100 Persian news documents each of which is associated with 5 summaries. Each summary consists of several sentences of the original text, selected by a human expert. Some sentences are slightly modified and are not, therefore, an exact copy of any original sentences. Documents are categorized into six categories such as political, economic and so on. The length of documents ranges from 4 to 156 sentences. Overall, it has about 2,500 sentences. 
\subsection{Extracting Features and Scaling}
All features introduced in section \ref{subsec:learning-FE} are calculated. Pre-processing, sentence and word tokenization, stop words removal, and part of speech tagging is performed using the Hazm library \cite{_Hazm.}. The majority of features have a range between zero and one. Other features are passed to a min-max scaler to transform into the same range. For the category feature which is nominal, the one-hot-encoding method applied and six flag features used instead. 
\subsection{Target Assignment}
\label{sec:expr-TA}
In assigning the target to a sentence, as mentioned in section (\ref{sec:learning-TA}), the goal is to assign a number between 0 and 1, with higher values as an indicator that the sentence is present in the majority of golden summaries. Because exact matching between sentences is not possible, to resolve the question of presence in a single golden summary such as $g$, we calculated the cosine similarity of the desired sentence with each sentence: $s_j\in g$ . Then the maximum value of these similarities is selected as an indicator of presence. This indicator is then calculated for other golden summaries and their average is assigned to the sentence as the target. 
\begin{equation}
Target(s)=\frac{\Sigma_{g_i\in G}^{}\max_{s_j\in g_i}cosine\_similarity(s,s_j)}{|G|}
\end{equation}
in which G is set of summaries written for the document containing s. This is an additional explicit evidence that target (and subsequently, ranking) is related to the document.
\subsection{Training Model}
A vast collection of scikit-learn tools were used for the learning phase. K-fold cross-validation is applied with k=4 and split size of 0.25. Three different regression methods were applied, including Linear Regression, Decision Tree Regression, and Epsilon-Support Vector Regression(SVR). Overall results were the same with minor differences. Thus only the SVR result is reported. Various values for parameters were examined but the best results were achieved by epsilon=0.01, kernel=rbf, and default values for other parameters. With the aim of evaluating summary qualities, the fitted regressor of each run was used to rank documents sentences in the test set. To compare with each standard summary, a summary with the same count of sentences was produced, and compared by ROUGE. Averaging these ROUGE scores over each document and then over the dataset, the overall quality of summaries produced by the model can be obtained.

The same process was repeated with a random regressor that needed no training, and which simply assigns a random number between zero and one to any given sample. Apart from measuring the performance of this regressor on the test set, the quality of summaries produced is evaluated and reported as a baseline. The juxtaposition of this baseline and our measured results will demonstrate how effective our feature set was and how intelligent our whole system worked.	
\section{Results and Discussion}
\label{sec:results}
In section (\ref{sec:eval}) MSE, R2 and ROUGE scores are remarked as evaluation measures. The results of our experiments are reported below in terms of these measures. For better comparison, we also ran another experiment in which the random regressor was used for ranking sentences and producing summaries. Table \ref{tab:table} shows and compares MSE and R2 reported from these experiments. The results show that in experiment 2, the mean squared error is reduced and the r2 score is increased. This means that using document-aware features leads to a more accurate learned model, proving our hypothesis about the relationship between document features and target ranks.

\begin{table}
 \caption{Quality of the regression model’s predictions on the test set.}
  \centering
  \begin{tabular}{lll}
    \toprule
          & MSE     & R2  \\
    \midrule
    Experiment 1 & 0.03448  & 0.12238     \\
    Experiment 2 & 0.03068 & 0.17576      \\
    Experiment 3, random regression     & 0.17112       & -3.39857  \\
    \bottomrule
  \end{tabular}
  \label{tab:table}
\end{table}

\begin{figure}
  \centering
  \includegraphics[width=0.7\linewidth]{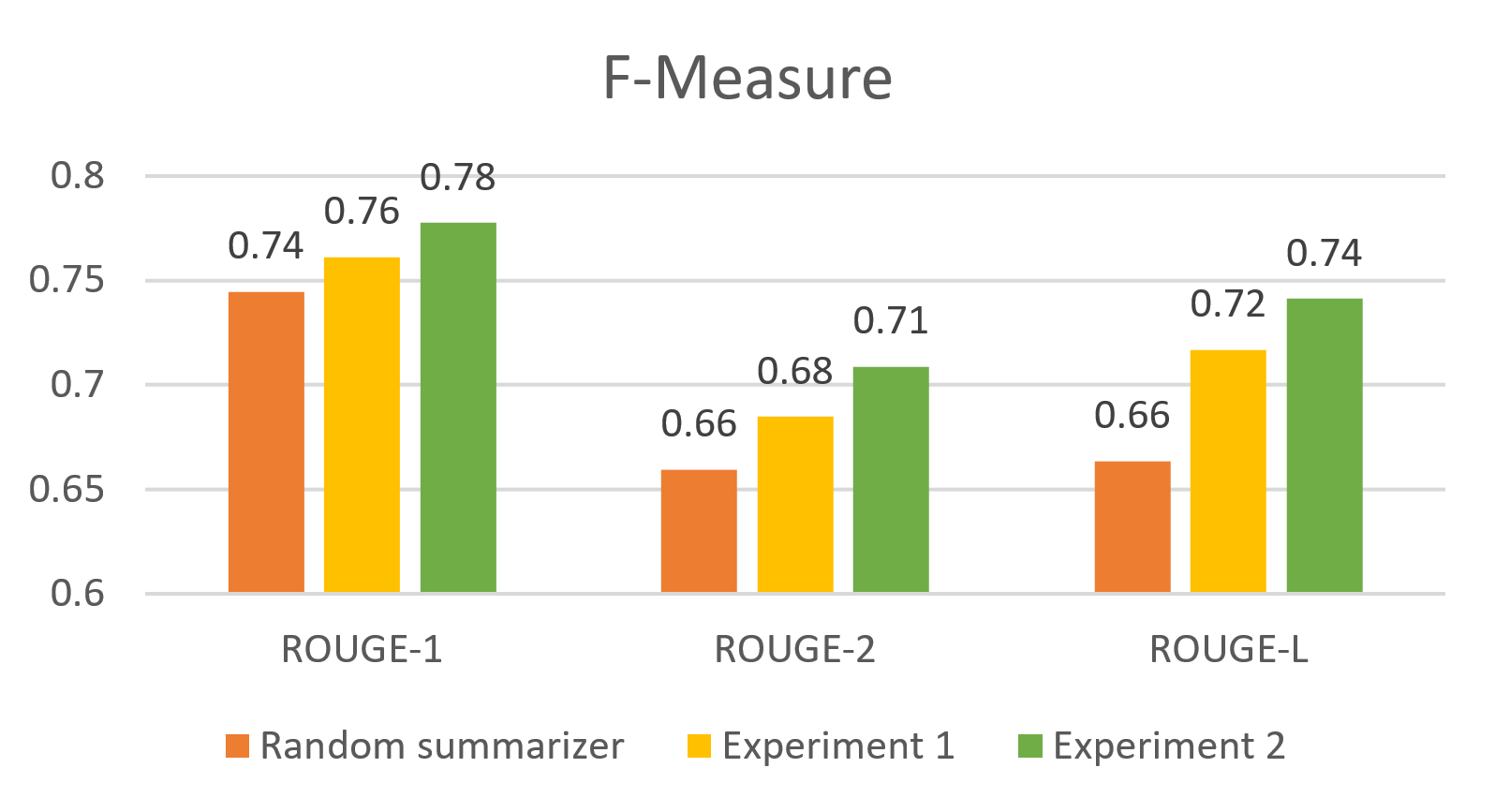}
  \caption{ROUGE Quality of produced summaries in terms of f-measure.}
  \label{fig:fmeasure}
\end{figure}

\begin{figure}
  \centering
  \includegraphics[width=0.7\linewidth]{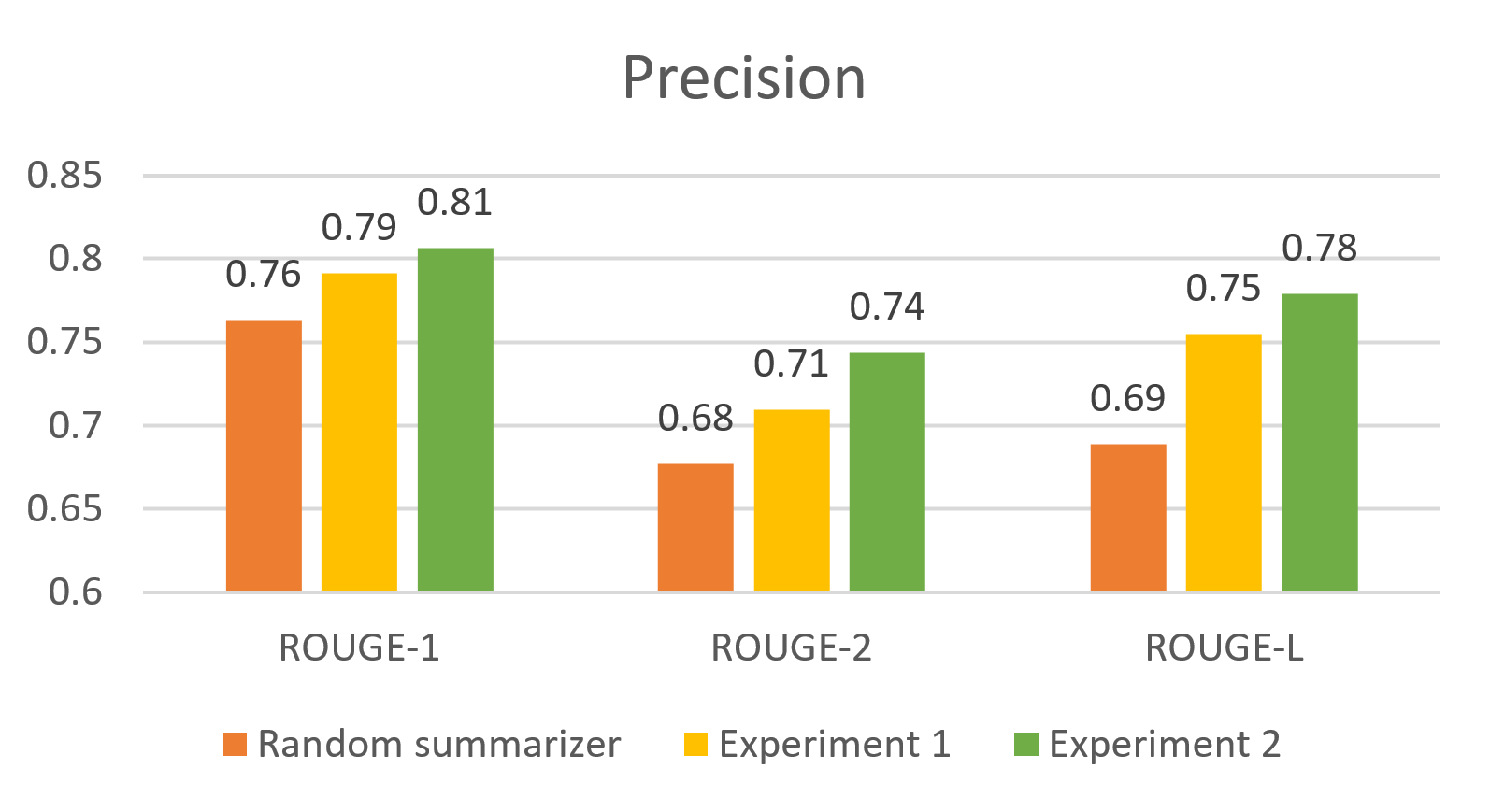}
  \caption{ROUGE Quality of produced summaries in term of precision.}
  \label{fig:precision}
\end{figure}

\begin{figure}
  \centering
  \includegraphics[width=0.7\linewidth]{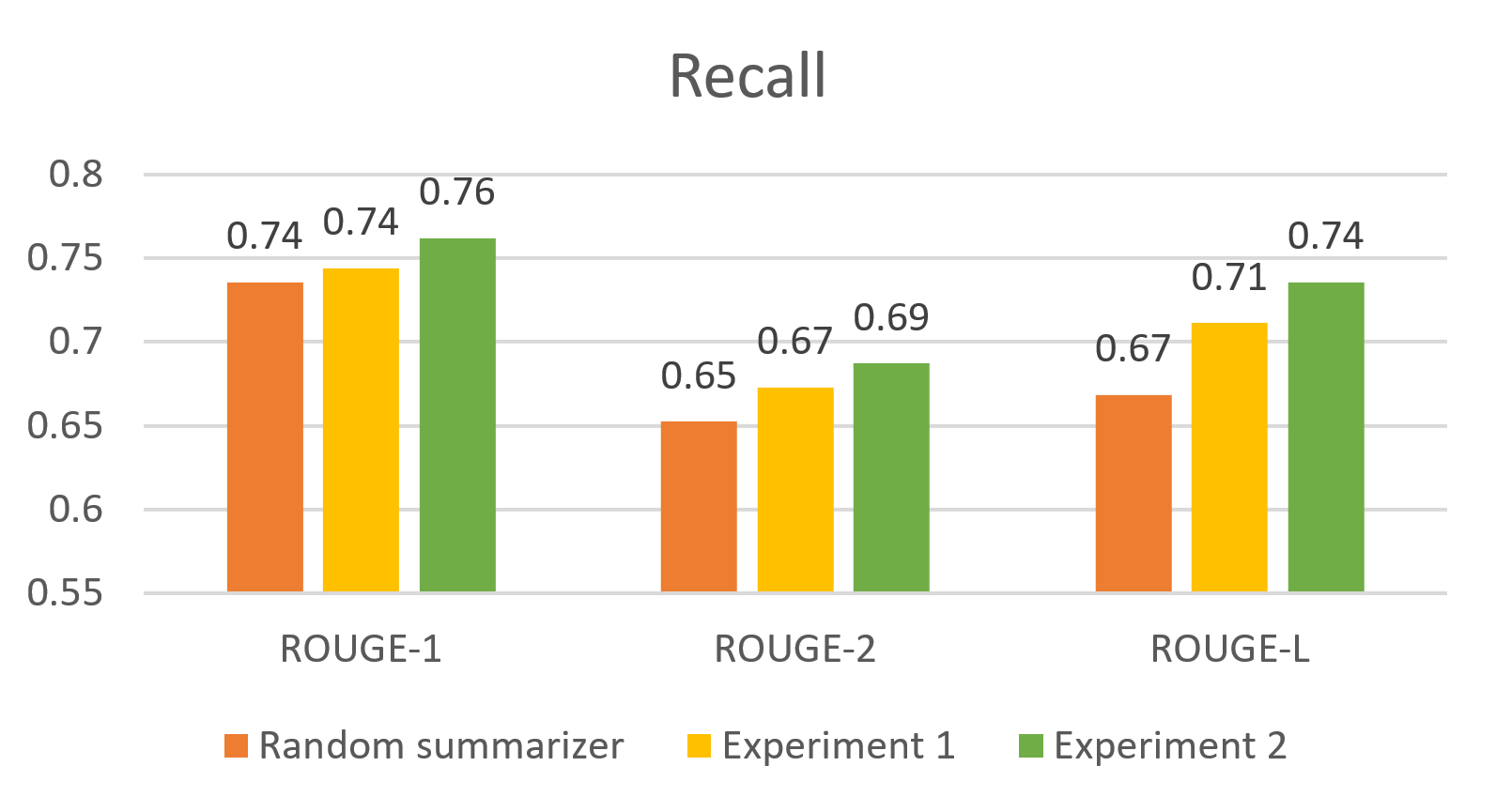}
  \caption{ROUGE Quality of produced summaries in term of recall.}
  \label{fig:recall}
\end{figure}

ROUGE scores are displayed separately in terms of precision, recall and f-measure in Figures \ref{fig:fmeasure} to \ref{fig:recall} respectively. F-measure scores are displayed in the figure \ref{fig:fmeasure}, comparing ROUGE-1, ROUGE-2 and ROUGE-L. Figures \ref{fig:precision} and \ref{fig:recall} allow comparison of precision and recall scores. The higher values gained in experiment 2, confirm that document-aware features perform better than unaware features.

These results are also interpretable from viewpoint of entropy-based decision tree methods. In learning phase, impurity of features within the whole dataset will be measured, and features having higher information gain will take place in upper levels of tree. But in summarization phase, within which decisions have to be made within a single document, impurity of those features may be low, causing less effective decisions and precision's. By incorporating document features, we help model to use different features (thus different trees) for different documents. 

Another insight gained from these charts is that a random summarizer resulted in scores more than 50\% in all measures, and without using document-aware features, the model achieves a small improvement over a random summarizer.

\section{Conclusion}
\label{sec:conclusion}
This paper has discussed that in supervised extractive summarization, we cannot learn to rank by considering dataset sentences as independent educational examples. The rank of sentences is dependent on each other within a document. To overcome this issue, we suggested incorporating document features explicitly in the feature vector of sentences. We also suggested using features that take into account the properties of document. We named this kind of features as document-aware. Conducted experiments demonstrated the benefit of adding explicit document features, as well as document-aware features, both in model precision and summary quality. For future work, more document-aware features can be examined. It is also possible to run the same experiments on an English (or any other language) dataset, if available. Another clue for study is measuring degree of entropy difference between dataset and single documents, in a standard dataset.

Our source code is hosted on GitHub\footnote{\url{https://github.com/Hrezaei/SummBot}} and is published for later reference, further experiments and reproducing results. A web interface\footnote{\url{http://parsisnlp.ir/summ/form}} and a Telegram bot\footnote{\url{https://t.me/Summ_bot}}  is also implemented as demo.

\bibliographystyle{unsrt}  
\bibliography{template}  

\end{document}